\documentclass[conference,hidelinks]{IEEEtran}

\IEEEoverridecommandlockouts
\usepackage{cite}
\usepackage{amsmath,amssymb,amsfonts}
\usepackage{algorithmic}
\usepackage{graphicx}
\usepackage{makecell}
\usepackage{textcomp}
\usepackage{xcolor}
\usepackage{titlesec}
\usepackage{multirow}
\usepackage{stmaryrd}

\usepackage{hyperref} 

\begin{document}
\bstctlcite{IEEEexample:BSTcontrol}

\title{Multi-Frequency Federated Learning for Human Activity Recognition Using Head-Worn Sensors\\}

\author{%
  \IEEEauthorblockN{%
    Dario Fenoglio\IEEEauthorrefmark{2}\textsuperscript{*}\href{https://orcid.org/0009-0002-9112-9522}{\includegraphics[scale=0.005]{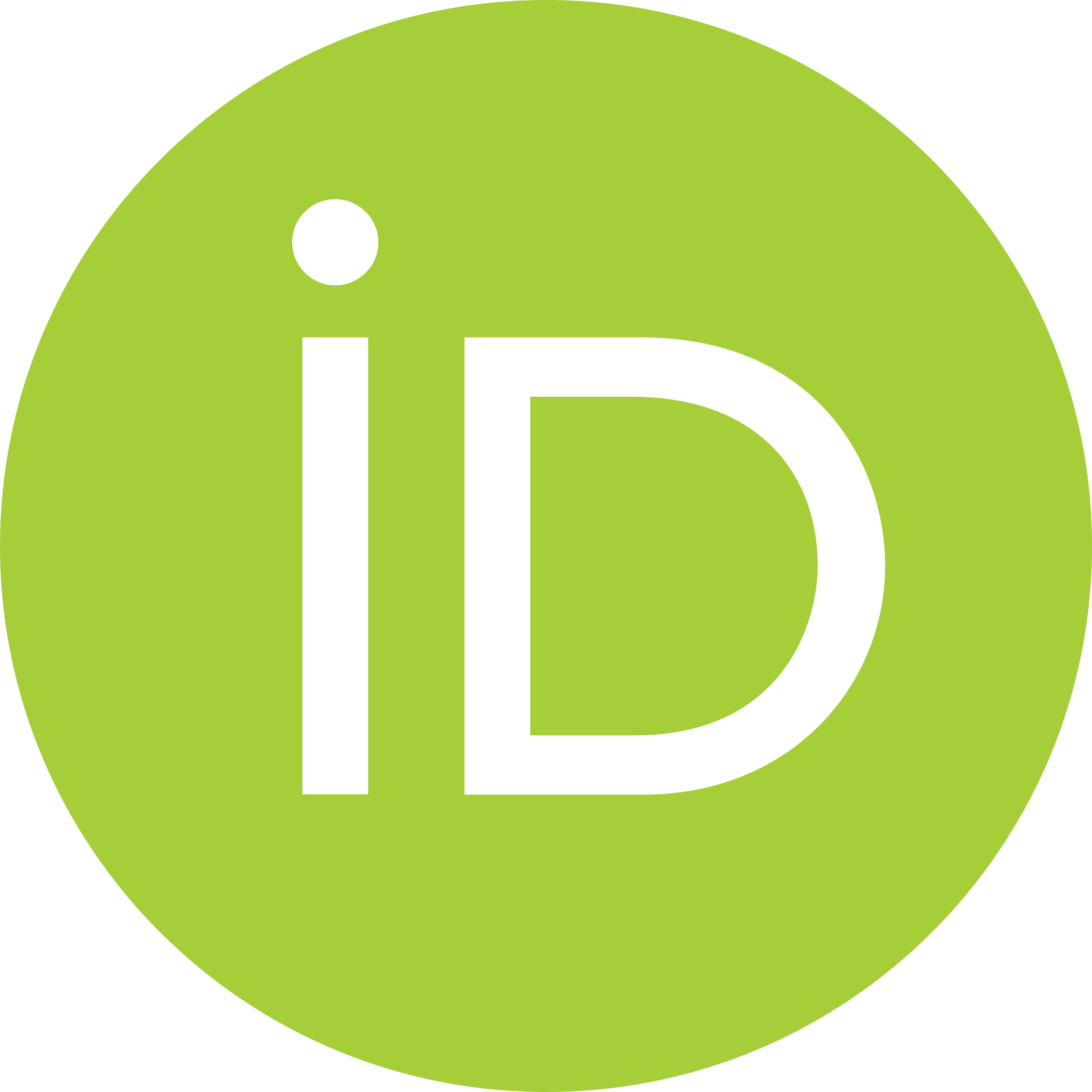}},
    Mohan Li\IEEEauthorrefmark{2}\textsuperscript{*}\href{https://orcid.org/0009-0005-3625-9801}{\includegraphics[scale=0.005]{figs/ORCID_iD.svg.png}},
    Davide Casnici\IEEEauthorrefmark{2}\href{https://orcid.org/0000-0003-0286-0728}{\includegraphics[scale=0.005]{figs/ORCID_iD.svg.png}},
    Matias Laporte\IEEEauthorrefmark{2}\href{https://orcid.org/0000-0001-7484-4116}{\includegraphics[scale=0.005]{figs/ORCID_iD.svg.png}},\\
    Shkurta Gashi\IEEEauthorrefmark{3}\href{https://orcid.org/0000-0001-6650-3784}{\includegraphics[scale=0.005]{figs/ORCID_iD.svg.png}},
    Silvia Santini\IEEEauthorrefmark{2}\href{https://orcid.org/0000-0002-0882-2004}{\includegraphics[scale=0.005]{figs/ORCID_iD.svg.png}},
    Martin Gjoreski\IEEEauthorrefmark{2}\href{https://orcid.org/0000-0002-1220-7418}{\includegraphics[scale=0.005]{figs/ORCID_iD.svg.png}},
    Marc Langheinrich\IEEEauthorrefmark{2}\href{https://orcid.org/0000-0002-8834-7388}{\includegraphics[scale=0.005]{figs/ORCID_iD.svg.png}}%
  }%
  \IEEEauthorblockA{\IEEEauthorrefmark{2} Università della Svizzera italiana, Switzerland}%
  \IEEEauthorblockA{\{dario.fenoglio, mohan.li\}@usi.ch}%
  \IEEEauthorblockA{\IEEEauthorrefmark{3} ETH Zurich, Switzerland}%
}

\maketitle
\begingroup\renewcommand\thefootnote{*}
\footnotetext{Equal contribution}
\endgroup
\begingroup
\renewcommand\thefootnote{**}
\footnotetext{\href{https://github.com/dariofenoglio98/Multi_frequency_STResNet.git}{https://github.com/dariofenoglio98/Multi\_frequency\_STResNet.git}}
\endgroup

\begin{abstract}
Human Activity Recognition (HAR) benefits various application domains, including health and elderly care. Traditional HAR involves constructing pipelines reliant on centralized user data, which can pose privacy concerns as they necessitate the uploading of user data to a centralized server. This work proposes multi-frequency Federated Learning (FL) to enable: (1) privacy-aware ML; (2) joint ML model learning across devices with varying sampling frequency. We focus on head-worn devices (e.g., earbuds and smart glasses), a relatively unexplored domain compared to traditional smartwatch- or smartphone-based HAR. Results have shown improvements on two datasets against frequency-specific approaches, indicating a promising future in the multi-frequency FL-HAR task. The proposed network's implementation is publicly available for further research and development.\textsuperscript{**}
\end{abstract}


\section{Introduction}
\label{sec:intro}

Human Activity Recognition (HAR) refers to the process of identifying and categorizing the specific activities performed by an individual through the analysis of various sensor data \cite{b0}. This technology has rapidly emerged as an essential tool with wide-reaching applications across numerous domains in recent decades. The importance of HAR lies in its ability to enable the provision of helpful context information that can be applied in various fields. Among these applications is the management of chronic diseases, where HAR can track and monitor patients’ physical activities to provide tailored healthcare solutions and interventions \cite{b1}. Similarly, in healthcare settings, detecting abnormal patient behavior can be automated through HAR, thus enhancing the efficiency of care and possibly even saving lives \cite{b2}. Furthermore, HAR offers insights into individual habits and routines, from personalized fitness tracking to occupational health and safety, allowing for the design of personalized programs to enhance overall well-being \cite{b3}. In summary, HAR not only adds a technological edge to many industries but also brings a human-centered approach to monitoring and understanding behavior. HAR applications are expanding and their impact on daily life is profound, shaping a new era of personalized, context-aware services and care.

Different approaches have been used in HAR, namely vision-based and wearable-based. Vision-based approaches utilize external sensors like cameras that provide a powerful way to recognize and analyze human activities by capturing visual data. However, they suffer from significant drawbacks. The efficacy of these tools is compromised when users are out of the sensing field. Privacy invasion is an even more critical concern, making them potentially unsuitable in personal or sensitive environments. These challenges underscore the limitations of vision-based HAR, particularly in the context of ubiquitous computing, where seamless and unobtrusive integration is a key consideration.

In contrast, wearable-based HAR offers a more flexible and privacy-respecting solution. 
Magnetometers, gyroscopes, and accelerometers –– essential components of inertial measurement units –– wearable sensors have become prominent in HAR for their ability to overcome some of the limitations of external sensors, such as their poor portability.
Being compact and easily integrated into everyday devices like earbuds or glasses, these sensors align seamlessly with the principles of ubiquitous computing. Their unobtrusive nature allows for continuous monitoring and data collection, permitting a more intuitive and user-centered approach to activity recognition \cite{b5}.

Unfortunately, the introduction of machine learning, specifically deep learning (DL), still complicates privacy preservation as these algorithms require data for training. Even when data are collected in a more privacy-preserving manner, transmitting it to a centralized server for model training cancels users’ exclusive ownership of the latter, opening up possibilities for data misuse and inadvertent exposure. 

In response to these challenges, Google introduced Federated Learning (FL) \cite{b6} in 2016, an innovative ML paradigm that enables neural network models to be trained across multiple decentralized devices or servers, each possessing its data samples locally. This approach significantly enhances data privacy and security by ensuring user data remains on their device, presenting a promising equilibrium between advancing HAR capabilities and maintaining robust user privacy protections.

An interesting aspect of FL is device heterogeneity, i.e., different devices can collaboratively train a model with a common goal (e.g., HAR). Recent HAR studies (such as FLAME \cite{c1}) have explored multi-device FL in synchronized setups, e.g., if a user wears earbuds and smart glasses simultaneously, FLAME could be used to train joint models across the synchronized devices.

Differently from the existing work, this study explores multi-frequency FL in asynchronous setups, i.e., users can have diverse active sensors (e.g., combinations of magnetometer, gyroscope, and accelerometer); these sensors can sample at various frequencies (e.g., some devices at 5Hz, others at 40Hz); and, these devices –– despite utilizing varying sensors and recording frequencies –– can collaboratively build a joint HAR model. Furthermore, we focus on FL for head-worn wearable devices, a relatively unexplored domain compared to traditional smartwatch or smartphone-based HAR.

To this end, this work makes the following contribution: \textit{a novel multi-frequency FL method for HAR.} The method is based on an existing end-to-end learning approach, Spectro-Temporal Residual Network (STResNet) \cite{c2}, that we adapted to work in a federated and multi-frequency setup. We compared the novel method on two datasets against centralized and frequency-specific models. The results show that our multi-frequency model allows the exploitation of all available data (i.e., all clients and sensors), thus outperforming frequency-specific models. In addition, our model demonstrated high flexibility and robustness, maintaining high performance while accepting a variable number of input sensors.


The rest of the paper is organized as follows. Section \ref{sec:related_work} introduces related works in FL and HAR. Section \ref{sec:dataset} provides details of the two datasets used in this work. Sections \ref{sec:method} and \ref{sec:result} present, respectively, the used methods and the experiments with the corresponding results. Section \ref{sec:discussion} discusses these results, while Section \ref{sec:conclusion} provides concluding remarks from them, as well as potential directions for future research.

\section{Related Work}
\label{sec:related_work}
This section provides an overview of the related research areas. We highlight two main differences between previous works and this study: (1) we focus on FL for head-worn wearable devices, a relatively unexplored domain compared to traditional smartwatch- or smartphone-based HAR; (2) to the best of our knowledge, this is the first study that explores a multi-frequency FL method for head-worn HAR in an asynchronous setup. The closest method to ours is FLAME, which utilizes synchronization (time alignment) across devices from the same user. Thus, FLAME is useful for scenarios where users simultaneously use multiple devices. On the other hand, in our proposed method, users need only one of the multiple devices (or sensors) to participate in the FL process.

\subsection{Federated Learning}
The FL community has been growing fast since it was introduced. As concluded from recent surveys \cite{b7,b8}, most of the research advances focus on core challenges, including reducing computing costs, tackling system or statistic heterogeneity, and enhancing privacy protection. Optimized communication and aggregation strategies \cite{b11,b12} have been proposed to relieve the computational burden without hurting the overall performance. Considering the participation of heterogeneous hardware, efforts have been made with adaptive task-assigning based on device capability, dropout of incapable devices, or tolerance as a more friendly approach \cite{b8}. To handle the typical non-independent non-identically distributed (non-IID) and unbalanced local data, one may resort to personalization through clustering \cite{b15} or model adaptation \cite{b16,ex1}. Even though the data are locally preserved, the vanilla model sharing and aggregation are exposed to malicious attacks such as data poisoning. Many related advanced works have followed Secure Aggregation \cite{b19} and Differential Privacy \cite{b20} as reliable solutions.
Besides HAR, FL has been proven successful in many other fields such as the Internet of Things \cite{b21}, healthcare \cite{ex1}, vehicular systems \cite{b25}, and recommender systems \cite{b26}.

\subsection{Human Activity Recognition}
A recent survey \cite{b28} has well captured recent advances in the HAR task with various sensing modalities. Frequently used signals include inertia, electrocardiogram (ECG), and vital signs such as respiration and temperature. In addition to the general HAR task \cite{b29}, other targets such as hand gesture recognition \cite{b32} are also highly related.
The unexploited unlabeled data have also gained much recent attention \cite{b34}. A large amount of data remains at the edges and is not applied for model training because labeling them is an overwhelming and knowledge-demanding task. More researchers now try to incorporate them in a semi-supervised or unsupervised manner to achieve better performance \cite{b35}.

\subsection{Federated Learning for Human Activity Recognition}

Even though current smart devices can collect billions of sensor samples every day with great potential to improve HAR performance, the cost of data transmission and the invasion of individual privacy are difficult to comprehend. Konstantin et al. \cite{b37} have made one of the earliest contributions that deploy the HAR learning task with the FL framework to tackle privacy issues. Tu et al. \cite{b38} proposed FedDL, where the center HAR model merges local updates based on a dynamic sharing scheme to speed up the convergence while maintaining high accuracy. Ouyang et al. \cite{b39} introduced ClusterFL, a similarity-aware FL approach to cluster different clients in a multitasking manner to achieve high model accuracy and low communication overhead for HAR applications. Xiao et al. \cite{b40} developed advanced feature extraction approaches from sensor data to improve the overall performance. Unsupervised learning and personalization have also been proven powerful as future directions \cite{b41}.

Despite the great success of FL on HAR, few of these stuides have explored wearable device data collected with earbuds and glasses, or in a multi-frequency setup. A recent-to-date work \cite{b43} practiced leveraging wearable smart glasses data to achieve personalized treatments and interventions for enhanced healthcare outcomes. Following their promising results, we explore multi-frequency FL for HAR. 

\section{Datasets}
\label{sec:dataset}
Our head-worn dataset consists of the \textit{USI-HEAR Dataset} \cite{b44} and the \textit{OCOsense Smart Glasses HAR Dataset} \cite{b43}. An overview is summarized in Table~\ref{dataset}.

\begin{table}[htbp]
\caption{Summary of Datasets.}
\begin{center}
\begin{tabular}{|c|c|c|}
\hline
\textbf{Dataset} & \textit{\textbf{USI-HEAR}} & \textit{\textbf{OCOsense}} \\ 
\hline
\textbf{Participants} & 30 & 24 \\
\hline
\textbf{Device} & eSense earbuds & OCOsense Smart Glasses \\
\hline
\textbf{Sensors} & \makecell{3-axis accelerometer\\ 3-axis gyroscope} & \makecell{3-axis accelerometer\\ 3-axis gyroscope\\3-axis magnetometer\\pressure sensor\\3-axis Euler virtual sensor} \\
\hline
\textbf{Activities} & \makecell{Speak and Walk\\Head Shaking\\Speaking\\Nodding\\Eating\\Walking\\Staying} & \makecell{Sitting\\Standing\\Laying\\Walking\\Transition\\Jogging\\Stair Climbing} \\
\hline
\end{tabular}
\label{dataset}
\end{center}
\end{table}

\subsection{USI-HEAR Dataset}
The \textit{USI-HEAR} Dataset was collected with the eSense earbuds developed by Nokia Bell Labs \cite{b46}. These earbuds consist of two Bluetooth-enabled units, each equipped with one microphone, while the left unit further houses one 6-axis Inertial Measurement Unit (IMU) sensor, comprising one 3-axis accelerometer and one 3-axis gyroscope.

Participants were provided with one left earbud (containing the IMU). They performed seven scripted activities, each lasting 3 minutes, with the data subsequently transferred to the experimenter’s laptop for verification. The experiment involved seven distinct activities, each carefully selected to represent a range of non-interacting and interacting behaviors. These activities were:

\begin{itemize}
    \item {\it \textbf{Speak and Walk}}: Participants combined walking and speaking, illustrating the complexity of simultaneous activities.
    \item {\it \textbf{Head Shaking}}: Participants moved their heads horizontally, with different intensities and intervals, representing a gesture of disagreement or denial.
    \item {\it \textbf{Speaking}}: Participants engaged in verbal communication with the experimenters, reflecting natural speech patterns and intonations.
    \item {\it \textbf{Nodding}}: Participants were instructed to nod their heads with different intensities and intervals, simulating a common gesture of agreement or acknowledgment.
    \item {\it \textbf{Eating}}: Participants consumed food, allowing for the observation of jaw movements and related motions.
    \item {\it \textbf{Walking}}: Participants walked at a comfortable pace, capturing the dynamics of regular locomotion.
    \item {\it \textbf{Staying}}: Participants remained still or seated, providing a baseline for motion detection.
\end{itemize}

Overall, the dataset comprises more than 10 hours of streaming data for each channel of both gyroscope and accelerometer, with a universal downsampled frequency of 50Hz.

\subsection{OCOsense Smart Glasses HAR Dataset}
The dataset was collected in 2022 by Emteq Labs using their \emph{\textit{OCOsense} Smart Glasses}~\cite{b47}. The device is equipped with one 3-axis accelerometer, one 3-axis gyroscope, one 3-axis magnetometer, one pressure sensor (barometer), and one 3-axis Euler virtual sensor to combine data from the accelerometer and gyroscope to provide the orientation of the glasses in three dimensions (yaw, pitch, roll).

24 participants were asked to perform the following activities while wearing the smart glasses:

\begin{itemize}
    \item {\it \textbf{Sitting}} (39.3\%) includes \emph{Sitting}, \emph{Sitting Still}, \emph{Sitting-looking around}, \emph{Sitting-using a PC}, and \emph{Sitting-using a phone}.
    \item {\it \textbf{Standing}} (27.3\%) includes \emph{Standing}, \emph{Standing Still}, \emph{Standing-looking around}, and \emph{Standing-using a phone}.
    \item {\it \textbf{Laying}} (18.3\%) includes \emph{On the back}, \emph{On the left side}, \emph{On the right side}, and \emph{On the stomach}.
    \item {\it \textbf{Walking}} (9.1\%) includes \emph{Walking}, \emph{Walking-looking around}, and \emph{Walking-using a phone}.
    \item {\it \textbf{Transition}} (2.2\%) includes \emph{Sitting down} and \emph{Standing up}.
    \item {\it \textbf{Jogging}} (1.7\%) includes \textit{Jogging}.
    \item {\it \textbf{Stair climbing}} (1.7\%) includes \textit{Stair climbing}.
\end{itemize}

The dataset contains 1.7M samples (9.5 hours duration in total) approximately equally distributed among all participants. The sampling frequency matches the earbuds dataset at 50Hz.

\section{Methods}
\label{sec:method}

In this section, we describe our comparisons among different centralized training models, to choose the best-performing model for FL. In addition, we describe the FL setup and the multi-frequency network.

\subsection{Centralized Machine Learning}
The DL pipeline employed in this study encompasses five distinct neural network architectures. Among these, four are 1D convolutional neural networks (ConvNets), and the remaining model is a deep multimodal spectro-temporal residual neural network (STResNet) \cite{b48}.

\vspace{-0.25em}
\begin{table}[htbp]
\caption{Comparison of Structures Among Four 1D ConvNets}
\begin{center}
\begin{tabular}{|c|c|c|c|c|}
\hline
\textbf{Model}&\textbf{\#Conv.}&\textbf{\#Dense}&\textbf{Act. functions}&\textbf{\#params.} \\
\cline{2-4}
\hline
ConvNet1&3&4 & LeakyReLU&78,680,443 \\
ConvNet2&3&3 & ReLU&5,962,453 \\
ConvNet3&2&2 & ReLU&1,937,575 \\
ConvNet4&2&2 & PReLU/ReLU&1,909,031 \\
\hline
\end{tabular}
\label{tab1}
\end{center}
\footnotesize Centralized model comparison among the four 1D convolutional neural networks (ConvNets) of structures, activation functions, and number of parameters.
\textbf{\#Conv.}: number of the convolutional layers.
\textbf{\#Dense}: number of the dense layers.
\textbf{Act. functions}: activation functions.
\textbf{\#params.}: number of the parameters.
\end{table}

The four 1D-convolutional deep models are characterized by different configurations of convolutional and dense layers, along with specific activation functions. The architectures are summarized in Table~\ref{tab1}. They share common features such as the softmax function as the final activation function and sparse categorical cross-entropy as the loss function. To mitigate overfitting, L2 regularization (rate = 0.0001) and dropout (rate = 0.5) were employed. Additionally, early stopping was implemented using the validation loss as the stopping criterion, further safeguarding against overfitting.

The STResNet model builds upon the concept of end-to-end unimodal time-series classification using residual networks. It incorporates multimodal and spectro-temporal information fusion, essential components of a successful HAR system. STResNet extracts channel-specific spectro-temporal information for each sensor channel. The spectral information is obtained by calculating the logarithm of the amplitude spectrogram in decibels for each input window. The temporal representation is extracted by residual blocks containing CNN layers with 1-dimensional (1D) filters. The shortcut connections in the residual blocks combat the gradient vanishing problem, making training more tractable. L2 regularization and dropout are applied to the dense layers, and the final output is provided by a softmax layer, representing class probability for each of the seven activities. Among the models finally employed in this study, STResNet stands out as the second most computationally demanding model, with a total of 14,005,415 trainable parameters. This substantial complexity is indicative of the model’s capacity to capture intricate patterns and relationships within the data. However, it is second only to ConvNet1 in terms of computational demand, reflecting a careful balance between model complexity and computational efficiency within the overall DL pipeline.

\subsection{Federated Learning}
In our study, we implemented the Weighted Federated Averaging algorithm via the Flower library~\cite{b50}. Each participating client performs one training epoch on their local dataset and then forwards their model weights and the count of their training samples to the server. To balance the contributions from all clients, the server applies a weighted average to these weights. We chose this approach due to the heterogeneous distribution of data across our datasets, resulting in varying numbers of samples per client. The model training was conducted using a sparse categorical cross-entropy loss function, combined with an Adam optimizer set at a learning rate of 0.0001. Considering the average number of samples per client, we standardized the batch size at 32 samples for both datasets. The final model was chosen based on its highest accuracy on the validation set during training.

For our FL environment, we exclusively employed the STResNet model, due to its superior performance in predicting activity on the \textit{USI-HEAR} dataset (as shown in \ref{subsection:A}). To ensure a precise comparison between centralized and federated learning, STResNet was implemented under both these settings across the \textit{OCOsense} and \textit{USI-HEAR} datasets. Additionally, to evaluate the robustness of FL under different conditions, we initially tested it with all available clients in the datasets, followed by a gradual reduction in the number of training clients. This allowed us to assess the impact of dataset size on the performance and effectiveness of the FL framework.

Specifically, to validate both centralized and federated approaches using the same test set clients, we employed a person-independent 5-fold cross-validation (ensuring non-overlapping test clients across different folds). This involved dividing the dataset into five distinct groups of clients, ensuring each client’s data was excluded once from the training process once. In each fold, clients not involved in the training were evenly split into validation and test sets. This strategy ensured a comprehensive and unbiased evaluation of both centralized and federated learning approaches.

\subsection{Multi-Frequency STResNet}
To address the challenge of clients equipped with sensors operating at varying frequencies, we developed a novel Multi-Frequency STResNet model. This model processes input signals from sensors with different sampling frequencies. For instance, we simulated a scenario in which half of the clients' sensors operated in a low-battery mode at 5Hz (or 3Hz), while the other half functioned at 40Hz. Instead of employing separate models for each battery mode — which would reduce the dataset size — we propose a versatile model that expands the potential data and client pool. Our approach involves creating distinct spectral-temporal encoders for each sensor (or channel), tailoring both temporal and spectral feature extraction to the sensor's sampling frequency.

\begin{figure}[htbp]
\centerline{\includegraphics[scale=0.45]{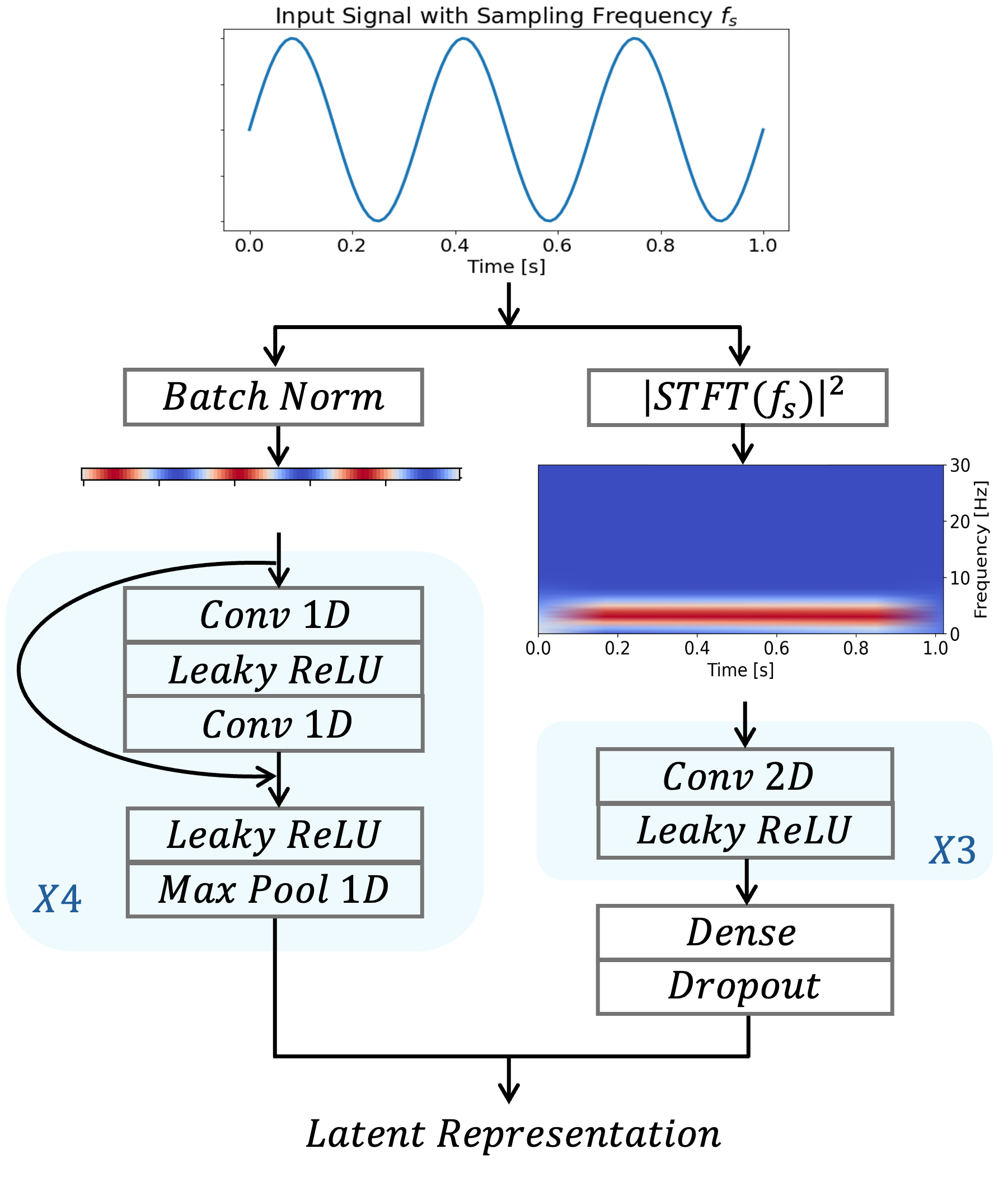}}
\caption{Encoder architecture for a single input channel. The input undergoes both a temporal and a spectral encoding to generate the latent representation.}
\label{fig0}
\end{figure}

As depicted in Figure~\ref{fig0}, our encoder receives a raw sensor signal as input and processes it through parallel pathways, which we call \textit{temporal} and \textit{spectral}. In the temporal pathway, after an initial batch normalization step, the signal undergoes four residual blocks. Each block consists of two 1D convolutions and two Leaky ReLU activations, followed by a single 1D max pooling operation to effectively capture the temporal dynamics. Concurrently, in the spectral pathway, the signal is transformed into a spectrogram and processed through three blocks, each containing a 2D convolution and a Leaky ReLU activation. To integrate the multidimensional spectral features, a dense layer and dropout regularization are employed.

\begin{figure}[htbp]
\centerline{\includegraphics[scale=0.35]{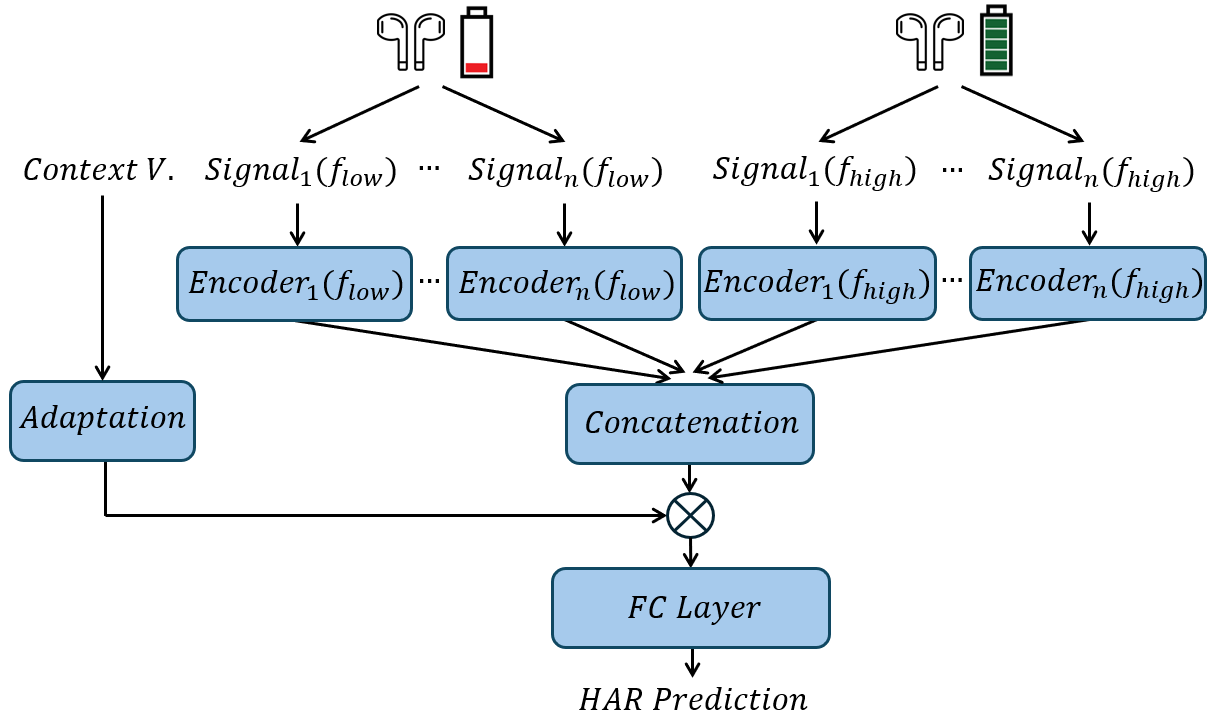}}
\caption{Multi-frequency model architecture designed for devices in low- and full-battery modes. A context vector is combined with the concatenation of the encoders' outputs to zero-out any absent sensors. Finally, fully connected layers produce HAR predictions from the combined latent representation.}
\label{fig1}
\end{figure}

Additionally, we introduced a context vector to mask activations from sensors that are not present (left-side of Figure~\ref{fig1}). This vector assigns a value of 1 where sensor input is available and 0 otherwise. Likewise, input channels are set to 0 in the absence of sensor data. Prior to the fully connected layers, the context vector is utilized to zero out activations from absent sensors, thereby preventing consideration of non-zero values due to the bias values involved in the neural networks' training.

The rest of Figure~\ref{fig1} illustrates how our Multi-Frequency STResNet accommodates users with sensors in both low- and full-battery modes. We evaluated our model in a centralized environment using person-independent 10-fold Monte Carlo cross-validation, ensuring consistent training and testing on the same client groups for each iteration. We benchmarked our model against those trained exclusively on clients with either 5Hz or 40Hz signals, a model trained on all clients at 5Hz (including downsampling those at 40Hz), and an ideal scenario where all clients operate at 40Hz (i.e., no low-battery mode). Furthermore, as our model is capable of being evaluated on test clients at both 5Hz and 40Hz, assessments were conducted under each condition to facilitate a fair comparison with frequency-specific models.

\section{Experiments and Results}
\label{sec:result}
This section outlines the experiments conducted and the corresponding results. In Subsection~\ref{subsection:A}, we evaluate five DL models across various sensor streams, identifying the STResNet model as the most accurate. Subsection~\ref{subsection:B} illustrates the comparable accuracy of federated and centralized learning, emphasizing FL's scalability and adaptability, even with varying numbers of training clients, while still ensuring user privacy.  Finally, Subsection~\ref{subsection:C} presents the results for the novel approach, demonstrating the Multi-Frequency STResNet model's efficiency in handling different sensor frequencies, outperforming frequency-specific setups. It should be noted that all models in our experiments were person-independent, ensuring non-overlapping training and testing client groups, which is crucial for the generalizability and applicability of our findings to real-world scenarios.

\subsection{Model Selection with Centralized Training}
\label{subsection:A}
We first carry out centralized training with all five DL models to select the one with the best performance. This model will then be used for the FL setup. We use different sensor streams for comprehensive validation from the \textit{USI-HEAR} dataset: accelerometer (ACC), gyroscope (GYR), magnitude (MAG), first-order derivatives (DER), and all combined (ALL). The results are shown in Table~\ref{tab3} with accuracy scores expressed as percentages, and standard deviations as percentage points.

\vspace{-0.25em}
\begin{table}[htbp]
\caption{Comparison of Results Among Centralized Models}
\begin{center}
\begin{tabular}{|c|c|c|c|c|c|}
\hline
\textbf{Model}&\textbf{ACC}&\textbf{GYR}&\textbf{MAG}&\textbf{DER}& \textbf{ALL} \\
\cline{2-4}
\hline
ConvNet1& \makecell{38.95\\±17.29} &\makecell{50.77\\±15.53} &\makecell{52.89\\±13.97} &\makecell{61.74\\±17.52} &\makecell{57.12\\±13.50} \\
\hline
ConvNet2& \textbf{\makecell{42.91\\±14.94}} &\makecell{56.02\\±14.54} &\makecell{55.12\\±13.83} &\makecell{62.23\\±17.17} &\makecell{65.49\\±12.11} \\
\hline
ConvNet3& \makecell{34.86\\±14.05} &\makecell{45.64\\±17.74} &\textbf{\makecell{58.32\\±13.05}} &\makecell{56.03\\±15.32} &\makecell{61.78\\±11.43} \\
\hline
ConvNet4& \makecell{20.01\\±5.63} &\makecell{43.16\\±18.75} &\makecell{36.23\\±10.07} &\makecell{38.33\\±16.31} &\makecell{40.55\\±13.25} \\
\hline
STResNet & \makecell{38.86\\±16.31} &\textbf{\makecell{57.13\\±13.31}} &\makecell{57.12\\±14.74} &\textbf{\makecell{62.55\\±11.91}} &\textbf{\makecell{69.22\\±11.78}} \\
\hline
\end{tabular}
\label{tab3}
\end{center}
\footnotesize Comparison of accuracy (with standard deviation) among five centralized models with different input settings, validated on the \textit{USI-HEAR} dataset. STResNet outperforms with inputs of all sensor streams combined. \textbf{ACC}: accelerometer.
\textbf{GYR}: gyroscope. 
\textbf{MAG}: magnitude. 
\textbf{DER}: first-order derivatives. 
\textbf{ALL}: all above combined. 
\end{table}

The results show that among the original sensor streams, the gyroscope emerges as the most pertinent. Within virtual sensor streams, the derivatives stand out as a better input for most models compared to the magnitudes. The highest performance is attained with the STResNet model, taking advantage of all the sensor streams and exploiting the spectral information of the signals, underscoring the importance of a comprehensive approach in sensor data analysis. This consistent performance across different scenarios justifies our choice of STResNet for the rest of the experimental setup.

\subsection{Comparison between Centralized and Federated Learning}
\label{subsection:B}
Table~\ref{tab4} presents the accuracy, F1-score, and cross-entropy loss for both centralized and federated learning techniques across the \textit{USI-HEAR} and \textit{OCOsense} datasets. These results underscore the efficacy of FL in handling diverse HAR datasets. Notably, the performance metrics in the FL setup were comparable to those in the centralized setup. This equivalence highlights FL's ability to effectively train a global model without the need for direct data sharing from clients. 

\vspace{-0.5em}
\begin{table}[htbp]
\scriptsize
\caption{Comparison Between Centralized and Federated Learning with All Participants}
\begin{center}
\setlength{\tabcolsep}{4.5pt}
\begin{tabular}{|c|c|c|c|c|}
\hline
\multicolumn{1}{|c|}{} & \multicolumn{2}{c|}{\textbf{Centralized}} & \multicolumn{2}{c|}{\textbf{Federated}} \\
\cline{2-5}
\textbf{} & \textbf{\textit{USI-HEAR}} & \textbf{\textit{OCOsense}} & \textbf{\textit{USI-HEAR}} & \textbf{\textit{OCOsense}} \\
\hline
\textbf{Accuracy} & \makecell{\textbf{70.0 ± 4.7}} & \makecell{84.9 ± 2.7} & \makecell{69.43 ± 4.14} & \makecell{\textbf{85.19 ± 1.99}} \\
\hline
\textbf{F1-Score} & \makecell{\textbf{70.2 ± 4.8}} & \makecell{\textbf{87.8 ± 1.7}} & \makecell{67.26 ± 4.02} & \makecell{87.72 ± 1.72} \\
\hline
\textbf{CE Loss} & \makecell{\textbf{1.052 ± 0.165}} & \makecell{\textbf{0.389 ± 0.083}} & \makecell{1.105 ± 0.242} & \makecell{0.444 ± 0.085} \\
\hline
\end{tabular}
\label{tab4}
\end{center}
\footnotesize Federated learning shows competitive performance to centralized learning with both \textit{USI-HEAR} and \textit{OCOsense} datasets.
\textbf{CE Loss}: cross-entropy loss.
\end{table}

Table~\ref{tab5} compares the performance of centralized and federated learning techniques with varying numbers of training clients. 
Following this comparison, it becomes evident that our FL approach demonstrates robustness under these conditions. The similarity in results between centralized and federated methods, even with different number of participants, indicates a high degree of scalability and adaptability in the FL approach. This suggests that FL can maintain consistent performance despite variations in the size of training data and the number of clients, crucial in real-world applications where client availability may vary.

\vspace{-0.25em}
\begin{table}[htbp]
\scriptsize
\caption{Comparison Between Centralized and Federated Learning with Various Participants Numbers}
\begin{center}
\begin{tabular}{|c|c|c|c|c|}
\hline
\multicolumn{1}{|c|}{} & \multicolumn{2}{c|}{\textbf{Centralized}} & \multicolumn{2}{c|}{\textbf{Federated}} \\
\cline{2-5}
\textbf{\#Part.} & \textbf{\textit{USI-HEAR}} & \textbf{\textit{OCOsense}} & \textbf{\textit{USI-HEAR}} & \textbf{\textit{OCOsense}} \\
\hline
2 & \makecell{55.55 ± 1.87} & \makecell{\textbf{77.14 ± 3.05}} & \makecell{\textbf{55.60 ± 2.43}} & \makecell{75.19 ± 1.97} \\
\hline
3 & \makecell{\textbf{58.41 ± 4.85}} & \makecell{\textbf{80.77 ± 2.15}} & \makecell{56.36 ± 3.87} & \makecell{79.18 ± 3.21} \\
\hline
4 & \makecell{\textbf{60.02 ± 4.22}} & \makecell{\textbf{81.92 ± 2.78}} & \makecell{58.64 ± 5.33} & \makecell{81.24 ± 2.47} \\
\hline
6 & \makecell{59.73 ± 7.72} & \makecell{84.43 ± 2.37} & \makecell{\textbf{60.34 ± 5.39}} & \makecell{\textbf{84.76 ± 1.76}} \\
\hline
8 & \makecell{\textbf{62.02 ± 5.13}} & \makecell{85.16 ± 2.01} & \makecell{61.53 ± 6.00} & \makecell{\textbf{86.60 ± 1.42}} \\
\hline
\end{tabular}
\label{tab5}
\end{center}
\footnotesize Comparison of F1-Score (with standard deviation) between Centralized and Federated Learning trained under different numbers of participants, with both \emph{\textit{USI-HEAR}} and \emph{\textit{OCOsense}} datasets. \textbf{\#Part.}: number of training participants.
\end{table}

\subsection{Multi-Frequency Model}
\label{subsection:C}
Table~\ref{tab6} provides a comprehensive comparison of F1-scores for the \textit{USI-HEAR} and \textit{OCOsense} datasets, across different numbers of participants included in the training (\emph{\#Part.} and their respective frequency) and diverse frequency settings. This table compares the performance of our Multi-Frequency STResNet model against various configurations as outlined in the first column: exclusively 5Hz clients, all clients downsampled to 5Hz (\emph{Down-5Hz}), exclusively 40Hz clients, and an ideal scenario with all clients at 40Hz (\emph{Ideal 40Hz}). Our multi-frequency model was trained on all the clients (both 5Hz and 40Hz) at their original frequency. To ensure a fair comparison, our our multi-frequency model (\emph{Multi-}), capable of processing both 5Hz and 40Hz frequencies, was tested under both these conditions (\emph{Multi-5Hz} and \emph{Multi-40Hz}) on the same test clients. 

Notably, our multi-frequency model exhibited superior F1-scores in both \emph{Multi-5Hz} and \emph{Multi-40Hz} configurations across both datasets, demonstrating its effectiveness over single-frequency settings (5Hz and 40Hz clients). This improvement is mainly attributed to the model's ability to utilize all available original data to train a unified model. However, it should also be noted that the \emph{Down-5Hz} model (i.e., an approach that first downsamples all the data to the lowest joint frequency (5Hz in this case) and then trains a model) slightly outperforms the multi-frequency approach. 

Precisely, in the \textit{USI-HEAR} dataset, the multi-frequency model achieved F1-scores of 63.26\% ± 3.08\% and 65.53\% ± 3.61\% for 5Hz and 40Hz, respectively. These scores significantly surpass the single-frequency models' scores of 59.30\% ± 3.33\% (5Hz) and 62.57\% ± 4.49\% (40Hz), and they closely align with the ideal outcome where all clients operate at a sampling frequency of 40Hz.

In the case of the \textit{OCOsense} dataset, a similar pattern emerges. Notably, as shown in Table~\ref{tab6}, downsampling to 5Hz resulted in a higher F1-score (86.17\% ± 2.29\%) than the ideal scenario of all 40Hz clients (85.77\% ± 2.17\%), suggesting that a 5Hz sampling rate is sufficient for accurate HAR tasks. Our multi-frequency model also performed better on the same test clients when downsampled to 5Hz compared to 40Hz. Once again, the multi-frequency model outperformed single-frequency setups, underscoring the versatility of our model in effectively handling diverse sensor frequencies. 
For a more detailed analysis, refer to the extended table in the appendix (Section~\ref{Appendix}), which includes both F1-scores and accuracy metrics. Additionally, the appendix presents analogous results for experiments conducted with a \emph{critical-battery} mode (i.e., a lower frequency of 3Hz).

\vspace{-0.25em}
\begin{table}[htbp]
\scriptsize
\caption{Comparison Between Configurations with Different Frequency}
\begin{center}
\begin{tabular}{|c|c|c|c|c|c|}
\hline
\multicolumn{2}{|c|}{\multirow{2}{*}{\textbf{Model}}} & \multicolumn{2}{c|}{\textbf{\textit{USI-HEAR}}} & \multicolumn{2}{c|}{\textbf{\textit{OCOsense}}} \\ \cline{3-6} 
\multicolumn{2}{|c|}{} & \textbf{F1-Score} & \textbf{\#Part.} & \textbf{F1-Score} & \textbf{\#Part.} \\ \hline
\multicolumn{2}{|c|}{5Hz} & 59.30 ± 3.33 & 7\textsubscript{5Hz} & 72.81 ± 14.17 & 5\textsubscript{5Hz} \\ \hline
\multicolumn{2}{|c|}{Down-5Hz} & 65.38 ± 2.39 & 14\textsubscript{5Hz} & 86.17 ± 2.29 & 10\textsubscript{5Hz} \\ \hline
\multicolumn{2}{|c|}{40Hz} & 62.57 ± 4.49 & 7\textsubscript{40Hz} & 79.74 ± 2.66 & 5\textsubscript{40Hz} \\ \hline
\multirow{2}{*}{\begin{tabular}[c]{@{}c@{}}Multi-\\(Ours)\end{tabular}} & 5Hz & 63.26 ± 3.08 & \multirow{2}{*}{\begin{tabular}[c]{@{}c@{}}7\textsubscript{5Hz}, 7\textsubscript{40Hz}\end{tabular}} & 85.38 ± 2.52 & \multirow{2}{*}{\begin{tabular}[c]{@{}c@{}}5\textsubscript{5Hz}, 5\textsubscript{40Hz}\end{tabular}} \\ \cline{2-2} \cline{3-3} \cline{5-5}
                       & 40Hz & 65.53 ± 3.61 &  & 83.45 ± 1.99 &  \\ \hline
\multicolumn{2}{|c|}{Ideal 40Hz} & 69.14 ± 2.96 & 14\textsubscript{40Hz} & 85.77 ± 2.17 & 10\textsubscript{40Hz} \\ \hline
\end{tabular}
\label{tab6}
\end{center}
\footnotesize Comparison of mean F1-Scores (with standard deviation) for different frequency configurations, with both \emph{\textit{USI-HEAR}} and \emph{\textit{OCOsense}} datasets, with half of the participants sampled at 5Hz and the other half at 40Hz. Our Multi-Frequency model is the only one that allows training with all clients at their original sampling frequencies. \textbf{\#Part.}: number of training participants.
\end{table}
\vspace{-0.5em}

\section{Discussion}
\label{sec:discussion}
We discuss the experimental results and our findings in this section, respectively, on the innovations in HAR modalities (Subsection~\ref{subsection:Novel}), insights into multi-frequency in HAR tasks (Subsection~\ref{subsection:multi}), and directions for future research (Subsection~\ref{subsection:challenge}).

\subsection{Novel Sensor Modalities in HAR}
\label{subsection:Novel}
Compared to traditional vision-based HAR, which demands image or video data from cameras, wearable-based HAR improves device availability and reduces data size with sensor streams while preserving competitive performance. As the main raw data source, inertial information is collected with IMUs, which can be fused with different devices and accessories. In this paper, we looked into datasets from earbuds and glasses, two relatively underexplored domains compared to other works, focusing on mobile and wrist-worn devices.

It should be noted that the on-body location of sensors could potentially introduce divergence in results: activities with subtle changes on face or head movements, such as speaking or eating, could be more distinguishable with sensor data above the neck. Likewise, activities with limb motion, such as running and walking, could be more clearly detected with sensors on the wrist or from the waist down. Predictably, a more systematic and comprehensive HAR framework should have multiple modalities to fully cover the range of human activities. Our work could be a good start to the concentration on more delicate differentiation of facial and head activities.

\subsection{Multi-Frequency HAR in The Wild}
\label{subsection:multi}
Recording frequency has a direct impact on the size of the data stream, storage requirements, and computational load. From the perspective of an end-user or client, we hope the model has an acceptable performance –– a generally faster response, higher device refresh rate, and less running-time memory taken –– even with minimum data frequency. From the perspective of a server or aggregator, accommodating a wider range of data frequency signifies a larger size of the training dataset, participation from more diverse users, and thus better performance for all. Furthermore, multi-frequency tolerance in FL introduces great possibilities for communication cost reduction and heterogeneous systems collaboration. We hope this work may pave the way for further research on this track.

In terms of multi-frequency in HAR tasks, we noticed that the \emph{Down-5Hz} model, i.e., an approach that first downsamples all the data to the lowest joint frequency (5Hz in this case) and then trains a model, has competitive results with the multi-frequency setup. This indicates a possible frequency threshold in the specific experimental datasets, where information introduced by a higher frequency is redundant. However, we expect that in real-life applications where the activities to be recognized are more dynamic, a 5Hz sampling rate would not be sufficient to achieve acceptable HAR performance. Thus, we hope our work inspires more flexible, lightweight, and energy-friendly frameworks.

\subsection{Challenges and Future Research}
\label{subsection:challenge}
Besides the future research directions above, there are more opportunities and challenges in the FL-HAR field. Related to this paper, we hope to address two important aspects in later works. First, we did not leverage the possibility of unlabeled data in this task. While millions of data streams appear in various sensor devices, end-users usually do not hold clean and well-labeled data. As our next step, we will explore both unsupervised and semi-supervised methods to further improve the model's quality and enlarge the training data size. Second, our work did not apply any personalization techniques, such as local re-training, which could be a solution to address the notorious non-IID problem in heterogeneous data. In the future, we may deploy a clustering or privacy-friendly knowledge-sharing approach to achieve better performance.

\section{Conclusion}
\label{sec:conclusion}
This paper has introduced a multi-frequency FL framework for the HAR task. Our framework builds one unified model to accommodate sensor data from different frequencies and classify human activities under various device status, such as low-battery mode. We tested our framework with simulated 5Hz and 40Hz data streams from two datasets (one collected from earbuds, another from smart glasses), obtaining promising results in recognizing human activities with a frequency-fusion approach over heterogeneous sensors. We hope our work brings attention to the data frequency issue in HAR, inspiring more research on heterogeneous-friendly FL systems.

\section*{Acknowledgment}
This study was funded by the projects TRUST-ME (205121L\_214991), SmartCHANGE (GA No. 101080965), and XAI-PAC (PZ00P2\_216405). Shkurta Gashi is supported by an ETH AI Center postdoctoral fellowship.

\bibliographystyle{IEEEtran}
\bibliography{IEEEabrv, bibliography}

\section{Appendix}
\label{Appendix}
This appendix provides additional results and metrics, complementing those presented in Section~\ref{subsection:C}. These results offer a broader view of the Multi-Frequency STResNet model's performance under various conditions and different metrics.

\subsection{Extended Performance Metrics}
Table~\ref{tab7} presents a comprehensive set of performance metrics for the Multi-Frequency STResNet model, introducing accuracy besides the F1-score. This table complements Table~\ref{tab6}, reported in the main results.

\vspace{-0.5em}
\begin{table}[htbp]
\tiny
\caption{Comparison of Accuracy, F1-Score, and Number of Training Participants (5-40Hz)}
\vspace{-2em}
\begin{center}
\setlength{\tabcolsep}{3.75pt}
\begin{tabular}{|c|c|c|c|c|c|c|c|}
\hline
\multicolumn{2}{|c|}{\multirow{2}{*}{\textbf{Model}}} & \multicolumn{3}{c|}{\textbf{\textit{USI-HEAR}}} & \multicolumn{3}{c|}{\textbf{\textit{OCOsense}}} \\ \cline{3-8} 
\multicolumn{2}{|c|}{} & \textbf{Accuracy} & \textbf{F1-Score} & \textbf{\#Part.} & \textbf{Accuracy} & \textbf{F1-Score} & \textbf{\#Part.} \\ \hline
\multicolumn{2}{|c|}{5Hz} & 60.17 ± 3.40 & 59.30 ± 3.33 & 7\textsubscript{5Hz} & 73.06 ± 6.43 & 72.81 ± 14.17 & 5\textsubscript{5Hz} \\ \hline
\multicolumn{2}{|c|}{Down-5Hz} & 65.74 ± 2.61 & 65.38 ± 2.39 & 14\textsubscript{5Hz} & 81.36 ± 2.44 & 86.17 ± 2.29 & 10\textsubscript{5Hz} \\ \hline
\multicolumn{2}{|c|}{40Hz} & 63.75 ± 4.05 & 62.57 ± 4.49 & 7\textsubscript{40Hz} & 76.90 ± 3.20 & 79.74 ± 2.66 & 5\textsubscript{40Hz} \\ \hline
\multirow{2}{*}{\begin{tabular}[c]{@{}c@{}}Multi-\\(Ours)\end{tabular}} & 5Hz & 63.69 ± 3.28 & 63.26 ± 3.08 & \multirow{2}{*}{\begin{tabular}[c]{@{}c@{}}7\textsubscript{5Hz}, 7\textsubscript{40Hz}\end{tabular}} & 80.00 ± 1.87 & 85.38 ± 2.52 & \multirow{2}{*}{\begin{tabular}[c]{@{}c@{}}5\textsubscript{5Hz}, 5\textsubscript{40Hz}\end{tabular}} \\ \cline{2-3} \cline{4-4} \cline{6-6} \cline{7-7}
                       & 40Hz & 66.34 ± 3.31 & 65.53 ± 3.61 &  & 78.78 ± 2.51 & 83.45 ± 1.99 &  \\ \hline
\multicolumn{2}{|c|}{Ideal 40Hz} & 69.73 ± 3.02 & 69.14 ± 2.96 & 14\textsubscript{40Hz} & 80.77 ± 2.47 & 85.77 ± 2.17 & 10\textsubscript{40Hz} \\ \hline
\end{tabular}
\label{tab7}
\end{center}
\end{table}
\vspace{-1em}

\subsection{Additional Experiments}
In addition to our primary experiments at a sampling frequency of 5Hz, we investigated the model's performance with signals sampled at a lower frequency of 3Hz. Table~\ref{tab8} presents the accuracy and F1-scores of our Multi-Frequency STResNet model, comparing its performance with configurations for exclusively 3Hz clients, clients downsampled to 3Hz (\emph{Down-3Hz}), exclusively 40Hz clients, and an ideal scenario where all clients operate at 40Hz (\emph{Ideal 40Hz}). Despite the reduced sampling rate, our multi-frequency model preserves its advantages over single-frequency models, registering an F1-score of 80.69\% ± 3.64\% and 80.49\% ± 2.51\% for 40Hz and 3Hz respectively, as compared to 78.90\% ± 3.01\% and 66.26\% ± 18.15\% for their single-frequency counterparts. Overall, it is observed that lowering the sampling frequency to 3Hz generally leads to a decline in HAR prediction accuracy. Nevertheless, particularly within the \textit{OCOsense} dataset, downsampling all clients to 3Hz yields results that are still comparable to using the full 40Hz frequency.

\vspace{-0.5em}
\begin{table}[htbp]
\tiny
\caption{Comparison of Accuracy, F1-Score, and Number of Training Participants (3-40Hz)}
\vspace{-2em}
\begin{center}
\setlength{\tabcolsep}{3.75pt}
\begin{tabular}{|c|c|c|c|c|c|c|c|}
\hline
\multicolumn{2}{|c|}{\multirow{2}{*}{\textbf{Model}}} & \multicolumn{3}{c|}{\textbf{\textit{USI-HEAR}}} & \multicolumn{3}{c|}{\textbf{\textit{OCOsense}}} \\ \cline{3-8} 
\multicolumn{2}{|c|}{} & \textbf{Accuracy} & \textbf{F1-Score} & \textbf{\#Part.} & \textbf{Accuracy} & \textbf{F1-Score} & \textbf{\#Part.} \\ \hline
\multicolumn{2}{|c|}{3Hz} & 57.48 ± 4.45 & 57.06 ± 4.24 & 7\textsubscript{3Hz} & 70.48 ± 6.34 & 67.05 ± 15.89 & 5\textsubscript{3Hz} \\ \hline
\multicolumn{2}{|c|}{Down-3Hz} & 63.68 ± 2.43 & 63.09 ± 2.46 & 14\textsubscript{3Hz} & 80.20 ± 3.33 & 82.95 ± 5.28 & 10\textsubscript{3Hz} \\ \hline
\multicolumn{2}{|c|}{40Hz} & 62.94 ± 3.65 & 62.05 ± 3.95 & 7\textsubscript{40Hz} & 76.40 ± 3.66 & 78.96 ± 2.22 & 5\textsubscript{40Hz} \\ \hline
\multirow{2}{*}{\begin{tabular}[c]{@{}c@{}}Multi-\\(Ours)\end{tabular}} & 3Hz & 61.48 ± 3.45 & 61.13 ± 3.52 & \multirow{2}{*}{\begin{tabular}[c]{@{}c@{}}7\textsubscript{3Hz}, 7\textsubscript{40Hz}\end{tabular}} & 78.48 ± 1.95 & 80.49 ± 2.51 & \multirow{2}{*}{\begin{tabular}[c]{@{}c@{}}5\textsubscript{3Hz}, 5\textsubscript{40Hz}\end{tabular}} \\ \cline{2-2} \cline{3-4} \cline{6-7}
                       & 40Hz & 64.85 ± 2.20 & 64.10 ± 2.03 &  & 77.44 ± 2.56 & 80.69 ± 3.64 &  \\ \hline
\multicolumn{2}{|c|}{Ideal 40Hz} & 68.58 ± 3.01 & 68.08 ± 3.09 & 14\textsubscript{40Hz} & 79.63 ± 2.20 & 82.86 ± 1.81 & 10\textsubscript{40Hz} \\ \hline
\end{tabular}
\label{tab8}
\end{center}
\end{table}

\end{document}